\pdfoutput=1

\documentclass[11pt]{article}

\usepackage[final]{acl}

\usepackage{times}
\usepackage{latexsym}

\usepackage[T1]{fontenc}

\usepackage[utf8]{inputenc}

\usepackage{microtype}

\usepackage{inconsolata}

\usepackage{graphicx}
\usepackage{graphicx} 
\usepackage{booktabs}
\usepackage{longtable}
\usepackage{url}
\usepackage{array}
\newcolumntype{C}[1]{>{\centering\arraybackslash}m{#1}}
\usepackage{pgfplots}
\usepackage{enumitem}

\title{\includegraphics[width=0.08\textwidth]{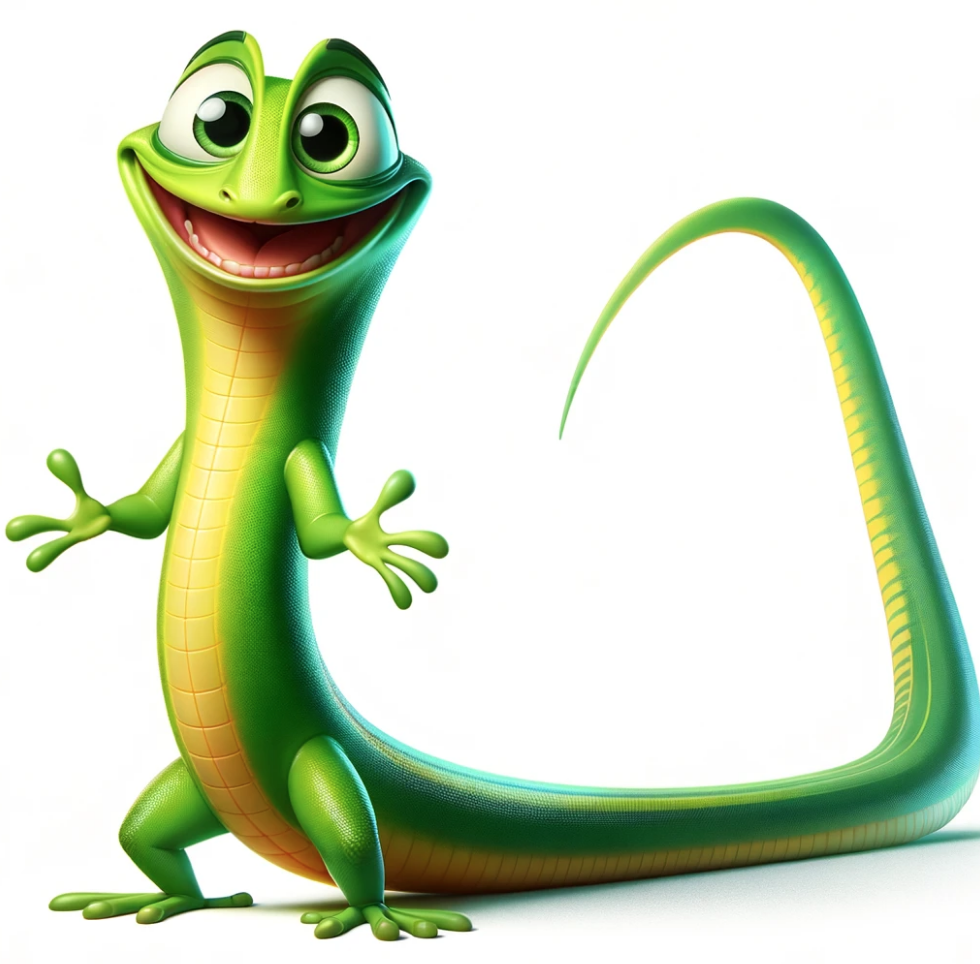} MLissard: Multilingual Long and Simple Sequential Reasoning Benchmarks}

\author{
  \textbf{Mirelle Bueno},
    \textbf{Roberto Lotufo}, 
    \textbf{Rodrigo Nogueira}
\\
\\
  School of Electrical and Computing Engineering, \\
  State University of Campinas (UNICAMP),
\\
\small{\url{m174909@dac.unicamp.br}}, 
\small{\url{{lotufo,rfn}@unicamp.br}}
}

\date{}

\begin{document}

\maketitle
\begin{abstract}

Language models are now capable of solving tasks that require dealing with long sequences consisting of hundreds of thousands of tokens. However, they often fail on tasks that require repetitive use of simple rules, even on sequences that are much shorter than those seen during training. For example, state-of-the-art LLMs can find common items in two lists with up to 20 items but fail when lists have 80 items. 
In this paper, we introduce MLissard, a multilingual benchmark designed to evaluate models' abilities to process and generate texts of varied lengths and offers a mechanism for controlling sequence complexity. 

Our evaluation of open-source and  proprietary models show a consistent decline in performance across all models and languages as the complexity of the sequence increases. Surprisingly, the use of in-context examples in languages other than English helps increase extrapolation performance significantly. The datasets and code are available at \url{https://github.com/unicamp-dl/Lissard}

\end{abstract}

\section{Introduction}
\begin{figure}[ht] 
    \begin{center}
        \includegraphics[width=\columnwidth]{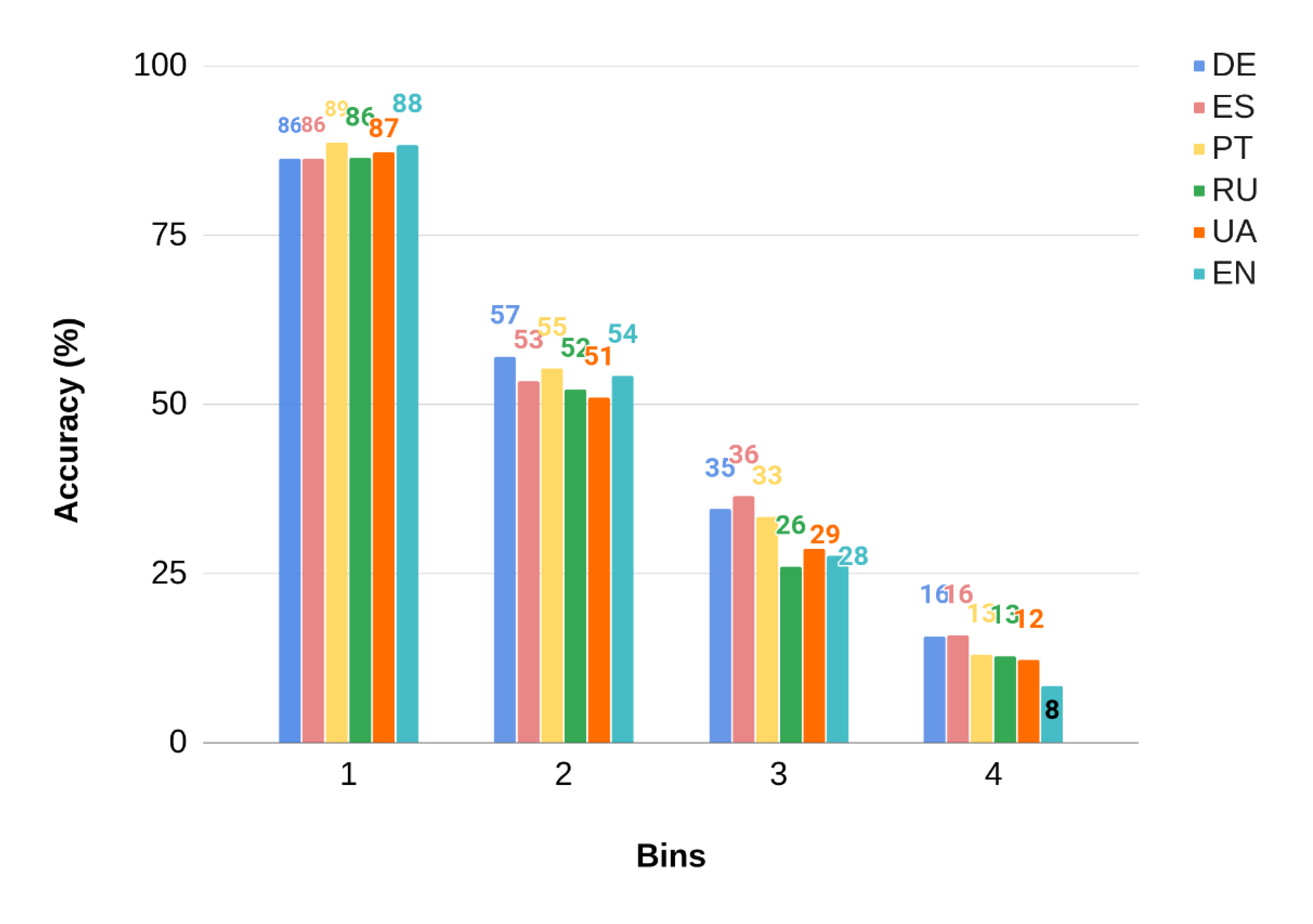}
    \end{center}
    \caption{Performance of GPT-4 on the MLissard benchmark. See Table~\ref{tab:key_entity_bins} for the definition of the bins.}
    \label{fig:gpt_4_all_tasks}
\end{figure}

The efficacy of language models, particularly in reasoning tasks, is significantly impacted by longer text lengths than those seen in training \citep{li2023functional, liu2023lost,lake2018generalization}. This phenomenon, referred to as ``Length Generalization'' or ``Length Extrapolation'' in the literature \cite{press2022train, zhao2023length}, is also common in models based on the Transformer architecture \cite{livska2018memorize,lewkowycz2022solving, delétang2023neural,zhou2023algorithms}. Notably, even Large Language Models (LLMs), known for their strong performance in a wide range of tasks and domains, are not immune to this problem \cite{anil2022exploring,chen2023extending}.

Recent research tried to address this challenge by modifications to the positional embeddings~\cite{press2022train, chi2022kerple,chi-etal-2023-dissecting, li2023functional,ke2021rethinking} or by using prompting strategies such as scratchpad~\cite{nye2021work} and chain-of-thought reasoning~\cite{wei2023chainofthought}. Nevertheless, there remains a lack of datasets specifically designed for the systematic evaluation of the problem.



While benchmarks such as ZeroSCROLLS \cite{shaham-etal-2023-zeroscrolls} and InfiniteBench\cite{zhang2024inftybench} were designed to evaluate models in natural language tasks that involve long sequences, its effectiveness in monitoring model performance degradation within the context of length generalization may be limited by lack of explicit control of task complexity with respect to sequence length. For example, when using natural language texts there is no guarantee that answering a question about a longer text is harder than responding to one about a shorter text. This limitation highlights the need for benchmarks that can explicitly manipulate and test the impact of sequence length on model performance. In benchmarks pertaining to dialogues \cite{longchat2023} and multi-document question answering \cite{liu2023lost}, techniques like retrieval-augmented generation (RAG) are prevalent, and therefore explicitly isolating the length extrapolation issue poses a challenge.

To address these aforementioned problems, we present MLissard, a multilingual benchmark that offer support for 6 languages (English, German, Portuguese, Russian, Spanish and Ukrainian) designed to evaluate the ability of models on tasks that require the use of repetitive simple rules, whose difficulty increases with respect to the sequence length. By incorporating varying degrees of difficulty within the same tasks, MLissard facilitates the identification of a models' breaking points. Given the syntactic nature of the datasets, researchers have the capability to generate new examples and increase the task difficulty, thus making it more challenging for newer and more capable models to be evaluated effectively.
This flexibility also mitigates the contamination problem -- where models may inadvertently be exposed to test datasets during their training \cite{ahuja2023mega,li2023task} -- since synthetic datasets can be generated as needed, a advantage over traditional, manually curated datasets.
At the time of this research, this is the first multilingual dataset designed to evaluate the quality of models in extrapolation via length.

Our analysis, which includes evaluations on proprietary models such as GPT-4~\cite{openai2023gpt4}, as well as open-source ones like Llama-3~\cite{dubey2024llama3herdmodels}, reveals a common trend among them. As illustrated in Figure~\ref{fig:gpt_4_all_tasks}, our findings underscore that irrespective of their architectures and parameter counts, all examined models demonstrate a performance degradation with increasing length, controlled by the number of key entities (see their definition in Table~\ref{tab:key_entity_bins}), required to solve the tasks. This indicates a common point of failure in generalization for LLMs, even for sequence lengths that are considerably shorter in terms of tokens than those seen during their pretraining or fine-tuning phases.


Our findings further demonstrated that the effect of extrapolation is not isolated; variables such as language and model size significantly influence the outcomes. For instance, despite English being a high-resource language, its performance was only average and was surpassed by other languages such as German. Moreover, ablation tests revealed improvements in extrapolation performance when in-context examples comprised a mixture of languages. This underscores the influence of language selection on the extrapolation capabilities of language models.


\section{Related Work}

The challenge of length extrapolation in the domain of natural language processing has been a persistent and long-standing issue. An array of studies has demonstrated that neural architectures encounter difficulties when confronted with sequences of longer than those they encountered during their training \cite{lake2018generalization,livska2018memorize,keysers2019measuring,dubois-etal-2020-location, nogueira2021investigating, welleck2022symbolic,lewkowycz2022solving, delétang2023neural,zhou2023algorithms}. Despite efforts to expand the context window in LLMs, this issue persists, particularly when tackling tasks involving complex reasoning~\cite{anil2022exploring}. 

Recent endeavors have been undertaken to enhance the  general performance of LLMs by employing prompt engineering techniques and by developing novel decoding methods aimed at expanding their capacity to extrapolate effectively over lengthy sequences of tokens.
For instance, \citeauthor{nye2021work} introduced the concept of a "scratchpad" that enables the model to generate draft responses in natural language before producing the final output. To assess the performance of this method, a range of tasks were employed, including math and coding tasks.
Moreover, studies by \citeauthor{ wei2023chainofthought} and ~\citeauthor{zhou2023leasttomost} demonstrated improvements by configuring the model to generate explanations for problem-solving and breaking down tasks into multiple interactive steps. These enhancements were particularly noticeable in tasks requiring the ability to extrapolate, such as SCAN~\cite{lake2018generalization} (compositional generalization), and mathematical reasoning.
Additionally, ~\citeauthor{bueno2022induced} showed that utilizing markups tokens as position representations help the model to generalize to longer sequences in tasks related to mathematical addition and compositional generalization. ~\citeauthor{han2023lminfinite} devised a decoding method to improve generalization over extended sequences.

In addition to techniques for customizing prompts, recent research has explored modifying the position encoding function of the original transformer architecture to enhance its extrapolation capabilities \cite{press2022train, chi2022kerple,chi-etal-2023-dissecting, li2023functional,qin2023exploring,chen2023extending}. For instance, ~\citeauthor{kazemnejad2023impact} conducted an evaluation of commonly used positional encoding methods, finding that omitting positional encoding altogether yielded superior results in downstream tasks.

The studies cited above illustrate multiple methods designed to address the challenge of extrapolation. Nevertheless, there is a notable gap in research concerning the development of diverse and standardized datasets specifically for assessing the generation and synthesis of extended text sequences by neural models. This gap is particularly notable given that many of the traditional datasets may already have been employed in the training of large language models.

\section{Datasets Description}



Our benchmark incorporates a combination of existing tasks, such as those from BIG-bench~\cite{srivastava2023beyond}, as well as newly developed ones. The criteria for selecting tasks were based on their ease of solution, the ability to expand new examples of varying lengths via scripting, and their effectiveness in exercising reasoning and memorization.

We intentionally excluded classical datasets (e.g., SCAN) from the analysis since their test sets are publicly available and many solutions have been extensively detailed in scientific literature, potentially making them familiar to large language models (LLMs).

In addition to English (EN), the language set includes German (DE), Spanish (ES), Portuguese (PT), Russian (RU), and Ukrainian (UA). We achieved this expansion by integrating automatic translation systems and using Python scripts to generate synthetic data.


The following sections describe the idea of key entities, tasks, and how evaluation was performed.

\subsection{Key entities}

The notion of key entities functions as an extrapolation factor within the context of a target task. For instance, in a task that seeks to identify common items between two lists, this extrapolation factor is defined by the number of items the model requires to analyze. Utilizing this factor allows for the augmentation of task complexity without modifying its properties. As a result, within specified ranges (bins), we can identify the model's breakpoints.

The choice of bins for each task was designed to reflect different difficulty levels: short, intermediate, long, and super long, for example, Bin 1 consists of sequences of shorter length, while Bin 4 comprises sequences of longer length. Table \ref{tab:key_entity_bins} describes the key entities and the respective lengths in each bin. The values defining the intervals of each bin vary for each task and were empirically determined, inspired by BIG-bench tasks.



\subsection{Tasks}

In total, four tasks were developed, and Table \ref{tab:summary_tasks} provides a summary of each one with input and output examples. Due to the high costs of paid APIs, we restricted our tests to 300 examples per task and language. To ensure balanced evaluations across different length partitions, we randomly selected 75 examples for each bin.

\begin{table*}
\centering
\begin{tabular}{p{0.20\linewidth}p{0.40\linewidth}p{0.20\linewidth}}
\toprule
\textbf{Task} & \textbf{Input Example}& \textbf{Output} \\
\midrule
Last Letter Concatenation & Abil Gaby & l y \\
\midrule  
Repeat Copy Logic  & Repeat 2 times school & school school \\
\midrule
Object Counting & I have a chair, and an apple. & 2 \\
\midrule        
List Intersection & A: abil,matt / B: matt, gaby & matt \\
\bottomrule
\end{tabular}
    \caption{Task Summary in the MLissard Benchmark.}
    \label{tab:summary_tasks} 
\end{table*}

\begin{table}
\centering
\begin{tabular}{p{0.1\linewidth}p{0.15\linewidth}p{0.1\linewidth}p{0.1\linewidth}p{0.1\linewidth}p{0.1\linewidth}}
    \toprule
    \textbf{Task} & \textbf{Key Entity}& \textbf{Bin 1}& \textbf{Bin 2}&\textbf{Bin 3}&\textbf{Bin 4}\\
    \midrule
    
    LLC & Names & 1-8 & 8-15 & 15-22 & 22-30 \\
    \midrule
    RCL &  Total Repetitions & 1-9&9-17&17-25&25-33 \\
    \midrule
    OC & Objects & 1-7&7-12&12-17&17-23 \\
    \midrule
    LI & Items: lists A and B & 1-46&46-91&91-136&136-181 \\
    \bottomrule
    \end{tabular}
    \caption{Key task entities: Last Letter Concatenation (LLC), Repeat Copy Logic (RCL), Object Counting (OC), and List Intersection (LI)  and their respective ranges in each bin in Figure~\ref{fig:gpt_4_all_tasks}.}
    \label{tab:key_entity_bins} 
\end{table}

\subsubsection{Object Counting}
The main goal of this task is to assess the proficiency in object counting within sequences, as shown in Table~\ref{tab:summary_tasks}. The input to the model is a sequence comprising a list of objects paired with their respective quantities and the expected output is a string with the total count of objects. Diverging from the original BIG-bench task that exclusively encompasses the enumeration of objects from predetermined categories like fruits, vegetables, or musical instruments, our method comprises object counting across different categories.

Automatic translation systems were used to generate the multilingual set, in this case, Google Translate. After this phase, a translation subset was selected for human analysis of the general quality of the translation.

    

\subsubsection{List Intersection}


The objective of this task is to find common items in two lists.
Items within the lists are composed of words from a designated target language, with both the words and their frequencies sourced from the FrequencyWords\footnote{\url{https://github.com/hermitdave/FrequencyWords/}}  repository. For each specific language, stop words and special characters were eliminated. Following this preprocessing phase, a random sampling of words was conducted.

The lists have equal sizes, but the number of overlapping items varies. The target output is the words in common, sorted alphabetically. If there are no items in common, "None" must be returned. 


\subsubsection{Last Letter Concatenation}


The Last Letter Concatenation task, as formulated in the Chain-of-Thought work \cite{wei2023chainofthought}, involves concatenating the last letter of each word within an input sequence comprised of random names. Table~\ref{tab:summary_tasks} provides an illustrative instance of the dataset, where the input sequence comprises randomly selected names obtained through the target language Name Census\footnote{Portuguese (PT) - \url{https://censo2010.ibge.gov.br/nomes/\#/ranking}

Spanish (ES) - \url{https://www.epdata.es/datos/nombres-apellidos-mas-frecuentes-espana-ine/373}

English (EN) - \url{https://www.ssa.gov/cgi-bin/popularnames.cgi}

German (DE) - \url{http://www.firstnamesgermany.com/}

Ukrainian (UA) - \url{https://census.name/ukrainian-name-database/}

Russsian (RU) - \url{https://census.name/russian-name-database/}}. 

In constructing our dataset, we applied a comparable methodology; however, we sampled the most common names from each target language and expanded the sample length to encompass sequences with an increase of up to thirty names. 






 \begin{figure}[ht] 
     \begin{center}
         \includegraphics[width=0.3\textwidth]{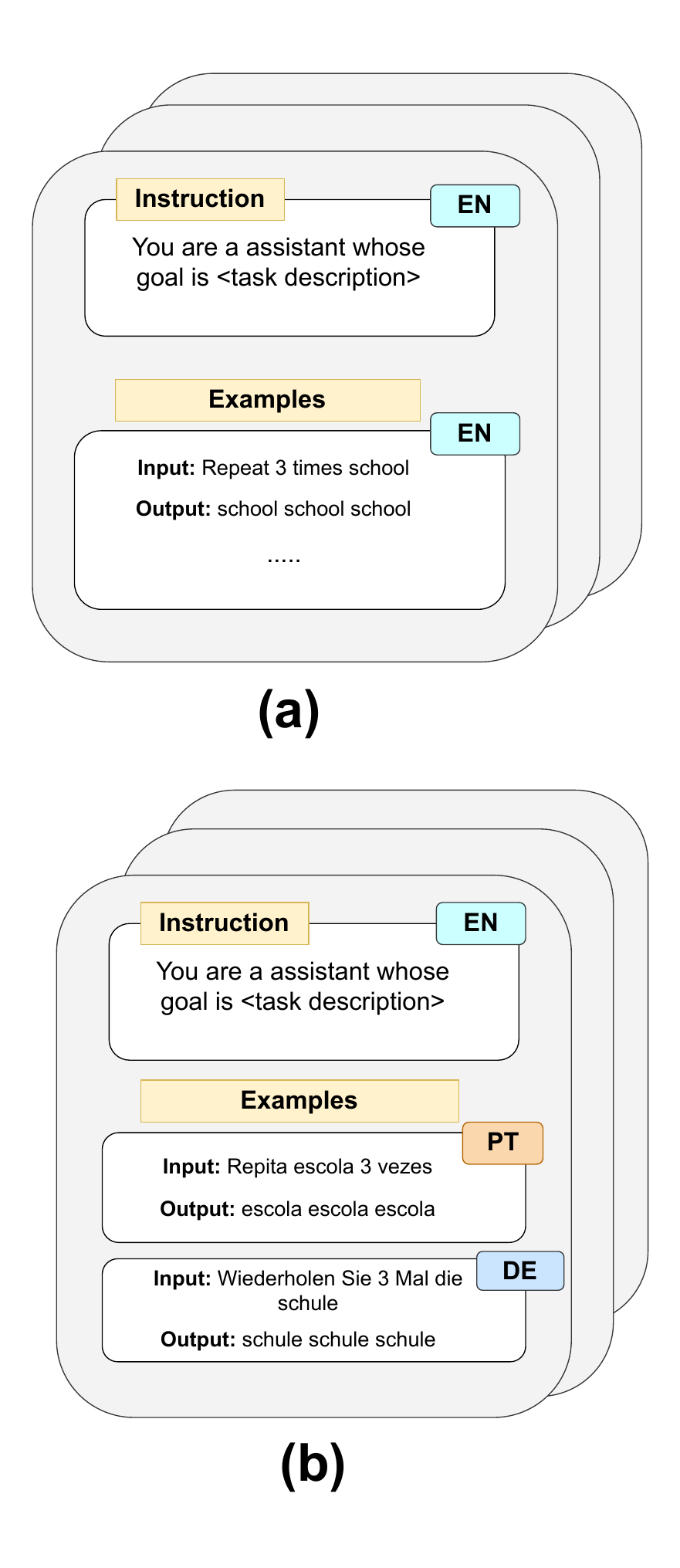} 
     \end{center}
     \caption{Template for evaluation. Being (a) Instruction and examples of tasks in the target language; (b) Instruction in the target language and multilingual examples.}
     \label{fig:mlissard_prompt}
 \end{figure}

\subsubsection{Repeat Copy Logic}
The task proposed by the BIG-bench evaluates language models' ability to comprehend and execute instructions involving repetitions, text-to-copy, basic logic, and conditionals, focusing on their extrapolation capabilities. 

\begin{figure*}[t] 
    \begin{center}
        \includegraphics[width=1\textwidth]{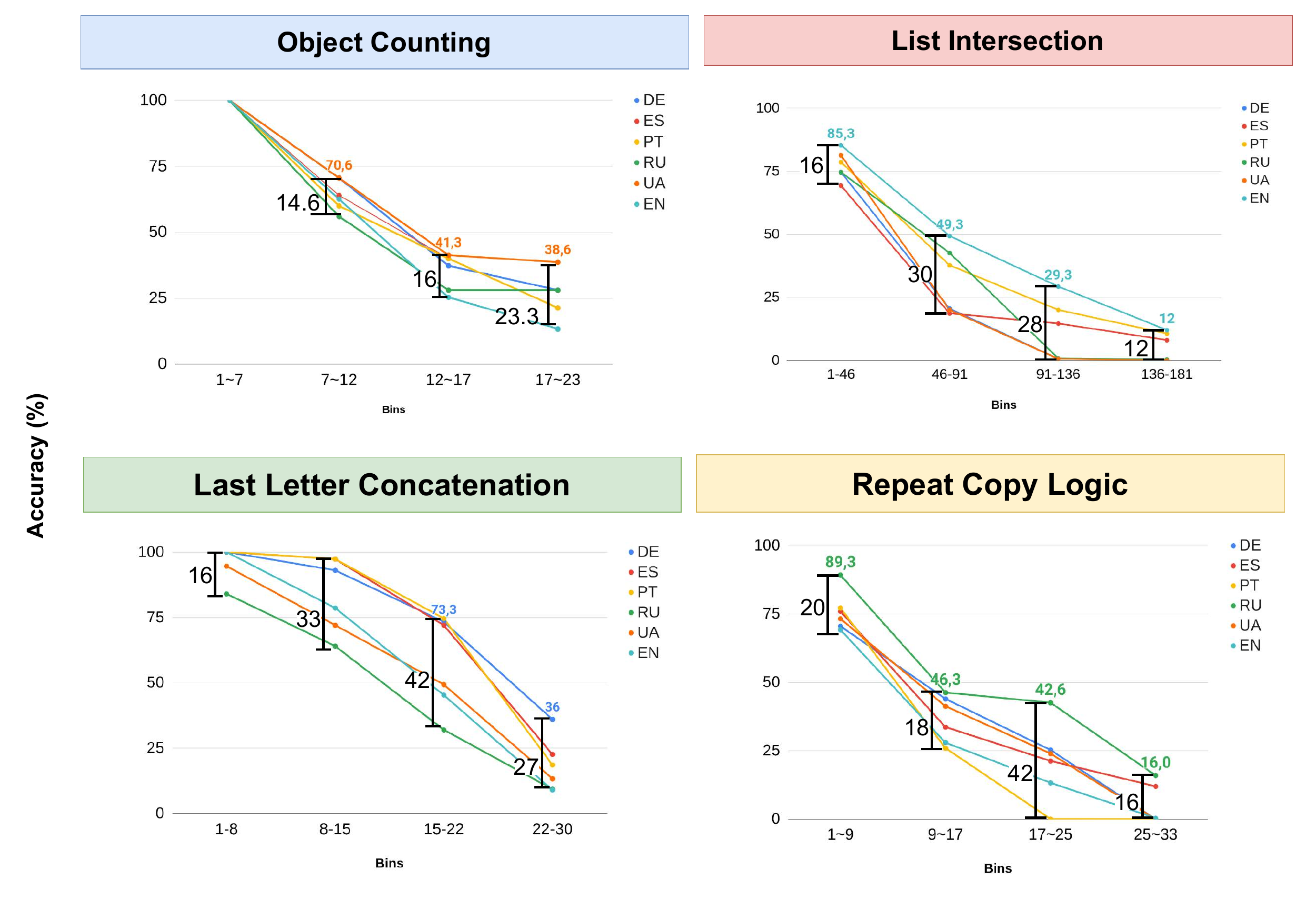} 
    \end{center}
    \caption{GPT-4 performance in the MLissard.}
    \label{fig:all_results}
\end{figure*}

\begin{figure*}[ht] 
    \begin{center}
        \includegraphics[width=1\textwidth]{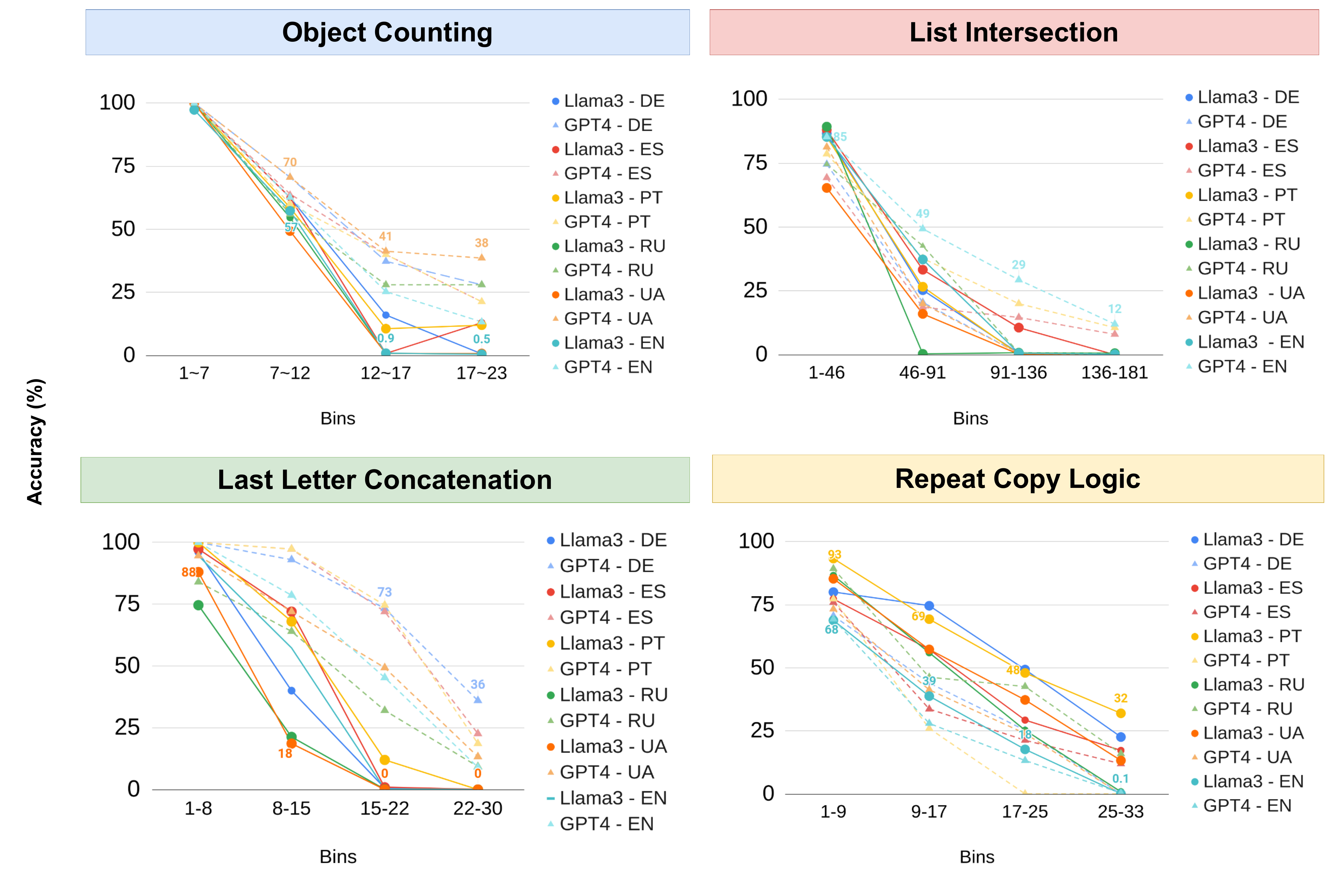} 
    \end{center}
    \caption{Comparison of Llama-3.1-405B vs. GPT-4 performance in the MLissard Benchmark}
    \label{fig:all_results_llama3_gpt_4}
\end{figure*}
\begin{table*}[ht]
\centering
\begin{tabular}{p{0.05\linewidth}|p{0.09\linewidth}p{0.09\linewidth}|p{0.09\linewidth}p{0.1\linewidth}|p{0.08\linewidth}p{0.1\linewidth}|p{0.1\linewidth}p{0.1\linewidth}}
\toprule
\textbf{Task} &\multicolumn{2}{c}{\textbf{Bin 1}} & \multicolumn{2}{c}{\textbf{Bin 2}}& \multicolumn{2}{c}{\textbf{Bin 3}} & \multicolumn{2}{c}{\textbf{Bin 4}}\\
\midrule
 & \textbf{Llama}& \textbf{GPT-4}& \textbf{Llama} & \textbf{GPT-4}&\textbf{Llama} & \textbf{GPT-4}&\textbf{Llama}&\textbf{GPT-4}\\
 \midrule
OC  & 100 & 100 & 58 & \textbf{63}&0.8 & \textbf{38}&0.7&\textbf{24.6}\\
LI  &\textbf{86} & 76& 26 & \textbf{29}&0.6 & \textbf{7.7}&0.1&\textbf{4}\\
LLC  & 95&\textbf{100}& 48.6& \textbf{85.8}&0.4&\textbf{ 60}&0&\textbf{16}\\
RCL  &\textbf{ 82}&73.3 & \textbf{57}&41.3 &\textbf{33}&24 &\textbf{15}&0.4\\
\midrule
AVG  &90.7&87& 47.4&54.7 &8.7&32.4 &3.9&11.7\\
\bottomrule
\end{tabular}
            \caption{Average accuracy of all languages per bin on tasks Object Counting (OC), List Intersection (LI), Last Letter Concatenation (LLC), and Repeat Copy Logic (RCL). Comparative result between the Llama-3.1-405B and GPT-4 models, highlighting in bold the best system performance in each bin.}
    \label{tab:llama-GPT_by_task} 
\end{table*}
\begin{figure*} 
    \begin{center}
        \includegraphics[width=1\textwidth]{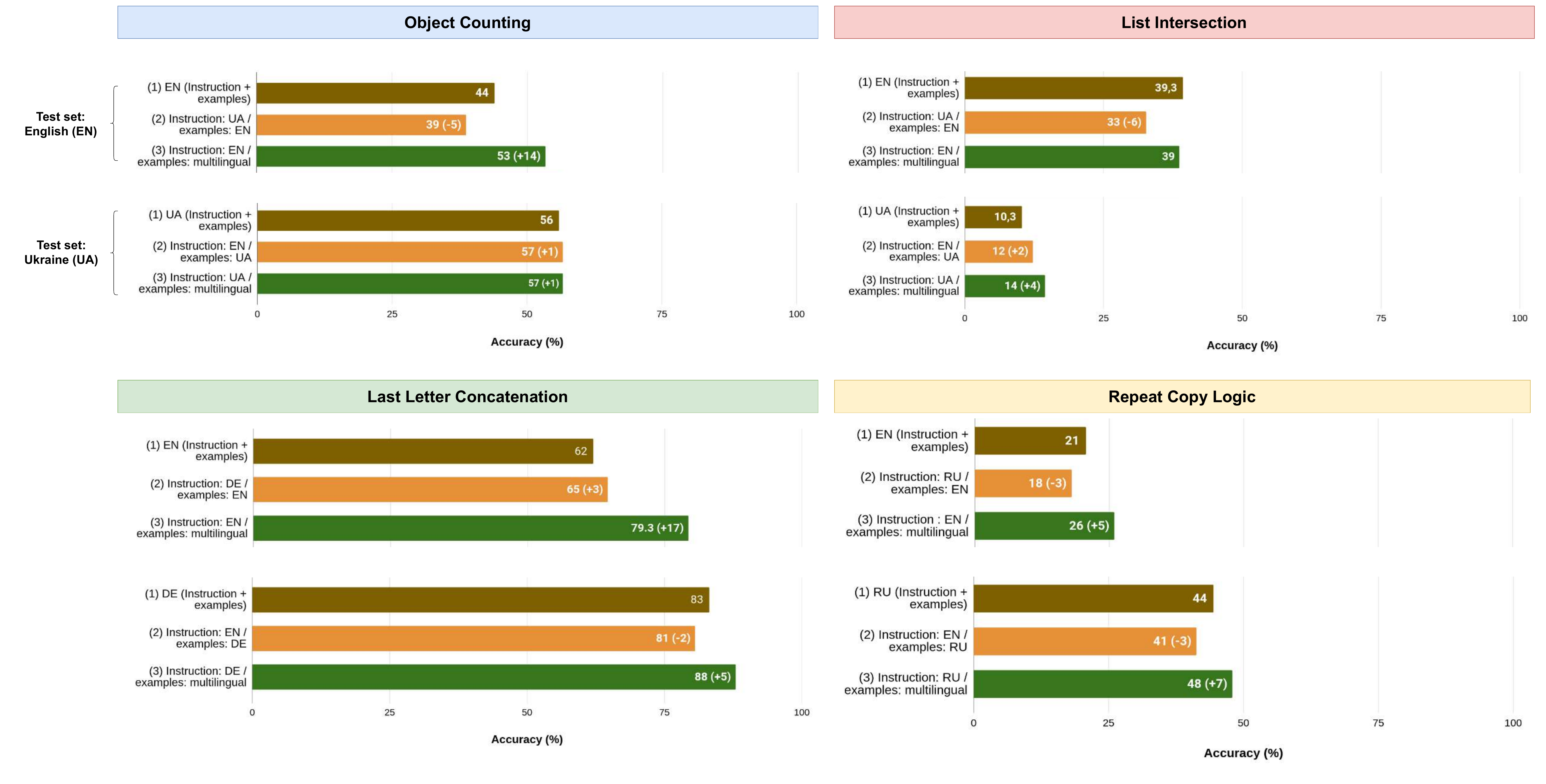} 
    \end{center}
    \caption{Average accuracy considering all bins. Since (1) Baseline - Both the instruction and the examples derive from the same target language; (2) instruction in the language that performed better or worse and a examples in the target language; (3) Instruction in target language and multilingual examples.}
    \label{fig:ablatio_results}
\end{figure*}

Our methodology for creating the dataset includes: i) Collecting responses to all input sequences from the BIG-bench repository\footnote{\url{https://github.com/google/BIG-bench/tree/main/bigbench/benchmark_tasks/repeat_copy_logic}}; ii) Filtering responses to retain only those correctly answered by GPT-4, which correctly answered 17 out of 32 original questions. We adopted this method to scale only the repetition factor; iii) Translating instructions using Google Translate and review the subset for accuracy; iv) Generate extrapolations on selected instructions, varying the repetition factor from 1 to 33 (see Table~\ref{tab:summary_tasks}). 

We randomly selected 15 of the 17 correctly answered questions for this phase.

\section{Baseline Methods}
The evaluation of each task involved analyzing responses from GPT-4 (gpt4-0613) and Llama-3 (Llama-3.1-405B-Instruct and Llama-3-instruction-70B) using greedy decoding. We observed no repetition issues. Each task was preceded by a predefined instruction (description of the task) with in-context examples: four for ``Object Counting,'' ``Find Intersection,'' and ``Last Letter Concat,'' and one for ``Repeat Copy Logic'' because inputs already provided sufficient information to perform the task. Both the instructions and examples were in the target language of the evaluation. For instance, English tasks used English instructions and examples (see Figure \ref{fig:mlissard_prompt} (a)). For the in-context examples used during model evaluation, we selected samples contained in the first bin, as these contain the smallest lengths. 

We utilized the exact match as the primary metric. This methodology is further modified in section~\ref{sec:ablation}, where we discuss the impact of cross-language inputs on model performance. 
\section{Results}


Figure~\ref{fig:all_results} presents the results obtained via GPT-4 in the target tasks and languages. Overall, there is a gradual decline in the performance of language models across tasks as complexity increases, as measured by the number of key entities in the input sequence. For instance, in the ``Object Counting'' task, when presented with inputs containing 1 to 7 objects, the model achieve approximately 100\% accuracy. However, their accuracy drops below 50\% when confronted with sequences with 12 to 17 objects. This behavior is reflected in the target languages as well, all of which present a loss of more than 50\% when dealing with more complex input sequences.


We also observed considerable variability in performance between languages depending on the specific task. For instance, differences ranging from 2.4 to 42 points are observed in the intermediate bins for tasks such as ``Last Letter Concatenation'' and ``Repeat Copy Logic''. These variations are intriguing as there doesn't appear to be a general language preference. For example, in the ``Last Letter Concatenation'' task, German, Portuguese, and Spanish outperform Russian by a margin of 42.6 points in the 15-22 bin. Conversely, in the ``Repeat Copy Logic'' task, Russian outperforms Portuguese by 42.5 points.


Contrary to the general trend observed in studies of multilingual models, English did not exhibit exceptional performance when compared to other languages. Except for the ``List Intersection'' task, English consistently remained at an average or lower accuracy level across bins.

Generalization performance also varies between tasks; as demonstrated in Table~\ref{tab:llama-GPT_by_task}, GPT-4 has greater difficulty executing the ``List Intersection'' and ``Repeat Copy Logic'' tasks. In the ``List Intersection'' task, the model achieves less than 10\% accuracy in bins 3 and 4. In the ``Repeat Copy Logic'' task, accuracy drops to below 25\% in the same bins. Both tasks require extensive memorization and state tracking. We hypothesize that these challenges, along with the increased sentence length, have influenced the observed performance outcomes.


Regarding the performance of open-source models in the MLissard benchmark, Figure~\ref{fig:all_results_llama3_gpt_4} illustrates that both models performed similarly in bin 1, with accuracy points ranging between 70 and 100. However, as task complexity increased from bin 2 onwards, differences in performance stood out. Except for the "Repeat Copy Logic" task, GPT-4 outperformed Llama-3.1-405B by 5 to 60 accuracy points (see Table ~\ref{tab:llama-GPT_by_task}).

On the other hand, in the ``Repeat Copy Logic'' task, there is a reverse comparison, where Llama-3.1-405B outperforms GPT-4 in all bins, with the difference ranging from 9 points to 16 points of accuracy.


In relation to language preference behavior, both the Llama-3.1-405B and GPT-4 models exhibit similar task-dependent variations. Llama-3.1-405B demonstrates more consistent performance across Portuguese, German, and English.


\subsection{Impact of model size}
\label{sec:ablation-model-size}
The Llama-3.1-405B model achieved state-of-the-art results in general NLP task benchmarks compared to the Llama-3-70B model. We investigated whether this performance trend is also evident in the MLissard benchmarks, especially in relation to the complexity indicated by the bins.


Table~\ref{tab:values} compares the average performance of each bin (for all MLissard tasks) using the Llama-3.1-405B and Llama-3-70B models. As expected, Llama-3.1-405B significantly outperforms Llama-3-70B across all languages and complexity bins. The largest differences between the models occur in bins 1 and 2, with performance gaps ranging from 16 to 43 points. In contrast, for bins 3 and 4, which involve more complex tasks, the performance improvement is less pronounced, with variations ranging from 0.3 to 11 points. This suggests that Llama-3.1-405B, like the 70B version, also struggles with long sequences.

\begin{table*}[ht]
\centering
\begin{tabular}{p{0.08\linewidth}|p{0.07\linewidth}p{0.1\linewidth}|p{0.07\linewidth}p{0.1\linewidth}|p{0.07\linewidth}p{0.1\linewidth}|p{0.07\linewidth}p{0.1\linewidth}}
\toprule
\textbf{Lang} &\multicolumn{2}{c}{\textbf{Bin 1}} & \multicolumn{2}{c}{\textbf{Bin 2}}& \multicolumn{2}{c}{\textbf{Bin 3}} & \multicolumn{2}{c}{\textbf{Bin 4}}\\
\midrule
 & \textbf{70B}& \textbf{405B}& \textbf{70B} & \textbf{405B}&\textbf{70B} & \textbf{405B}&\textbf{70B}&\textbf{405B}\\
 \midrule
EN  & 70.6 & 90  & 18.6 &48&0.1 & 0.7 &0&0.1 \\
PT  & 79.3& 96.6 & 24 & 63.3 &0.1 & 11.3&0&6 \\
ES  & 74&92.6 & 16.6&60 &0.1& 5.7 &0&6.5 \\
DE  & 74.6&91.3& 16.8 & 51.3 &0.5 &8.3 &0&0.3 \\
RU  & 60.6 & 88& 12.2& 38&0 & 0.8 &0&0.6 \\
UA  & 55.3& 86.6& 10.7 & 33.9&0.1 & 0.5&0&0.4\\
\bottomrule
\end{tabular}
    \caption{Average accuracy across all MLissard tasks was compared between the Llama-3-70B and Llama-3.1-405B models.}
    \label{tab:values} 
\end{table*}


\subsection{Can cross language improve extrapolation performance?}
\label{sec:ablation}

We aim to examine the impact on extrapolation performance by focusing on two components: 1) providing instructions in a different language than the target language, and 2) using mixed-language few-shot examples (see Figure~\ref{fig:mlissard_prompt} - (b)). For in-context examples, we used Portuguese, German, Ukrainian, and English. For the "Repeat Copy Logic" task, we provided two contextualized examples (English and Ukrainian), while for the other tasks, we provided four examples.


We conducted ablation tests on all tasks in the MLissard dataset using the GPT-4 model. For comparative purposes, we focused on the languages that achieved the highest and lowest performance in each task. We then compared these results with the baseline (both instructions and examples in the same language).


Figure~\ref{fig:ablatio_results} presents the experimental results for each task. As shown in the results, when we gave prompts in a language different from the test set, accuracy declined by an average of 2.3 percentage points. However, when we kept instructions in the test target language but included paraphrased examples contextualized in multiple languages, performance improved by an average of 6.25 percentage points. This improvement ranged from 2 points in the "List Intersection" task to 17 points in the "Last Letter Concatenation" task and remained consistent across all evaluated languages. These findings indicate that contextual examples in multiple languages can improve the quality of extrapolation.

\section{Conclusion}

We presented a multilingual benchmark to evaluate the ability of language models to deal with long texts across languages. Our approach distinguishes itself from existing benchmarks through the introduction of a control mechanism, which we refer to as "key entities." This mechanism enables us to systematically increase task complexity in tandem with sequence length. Furthermore, the ability to solve these tasks is predicated on the repeated application of simple rules, providing more control and enabling a detailed analysis of model performance in relation to the frequency of rule application. This contrasts with benchmarks that rely on lengthy natural language texts, where the relationship between text length and task difficulty may become obscured. Despite the apparent simplicity of these tasks, they reveal significant limitations in state-of-the-art LLMs concerning the processing and generation of text as lengths increase.  Our findings indicate that language and model size significantly affect extrapolation results. Moreover, including in-context examples in multiple languages improves MLissard's generalization performance.

\section{Limitations}
Our evaluations were conducted on a set of six languages, therefore,  the findings of this work may not necessarily extend to other languages, particularly low-resource ones. Additionally, we solely employed a standard prompt style for our evaluations, and the performance with more sophisticated techniques, such as chain-of-thought (CoT) prompting, remains to be investigated. Finally, given the limitation of our study to two models (GPT-4 and Llama-3), the results may not generalize to other LLMs.


\bibliography{custom}
\appendix

\end{document}